\documentclass{article} 
\usepackage{iclr2023_conference,times}

\usepackage{amsmath,amsfonts,bm}

\def\eqref#1{equation~\ref{#1}}

\def\1{\bm{1}}

\DeclareMathAlphabet{\mathsfit}{\encodingdefault}{\sfdefault}{m}{sl}
\SetMathAlphabet{\mathsfit}{bold}{\encodingdefault}{\sfdefault}{bx}{n}

\usepackage{hyperref}
\usepackage{url}
\usepackage{graphicx,subfig}
\usepackage{booktabs}
\usepackage{multirow}

\newtheorem{theorem}{Theorem}[section]
\newtheorem{lemma}[theorem]{Lemma}
\newtheorem{corollary}[theorem]{Corollary}

\title{Is Channel Independent strategy optimal for Time Series Forecasting?}

\iclrfinalcopy

\author{Yuan Peiwen , Zhu Changsheng \thanks{Corresponding author. Email: \texttt{zhucs@lut.edu.cn}} \\
School of Computer and Communication\\
Lanzhou University of Technology\\
Lanzhou, China \\
\texttt{\{yuanpw,zhucs\}@lut.edu.cn} 
}

\begin{document}

\maketitle

\begin{abstract}
There has been an emergence of various models for long-term time series forecasting. Recent studies have demonstrated that a single linear layer, using Channel Dependent (CD) or Channel Independent (CI) modeling, can even outperform a large number of sophisticated models. However, current research primarily considers CD and CI as two complementary yet mutually exclusive approaches, unable to harness these two extremes simultaneously. And it is also a challenging issue that both CD and CI are static strategies that cannot be determined to be optimal for a specific dataset without extensive experiments. In this paper, we reconsider whether the current CI strategy is the best solution for time series forecasting. First, we propose a simple yet effective strategy called CSC, which stands for $\mathbf{C}$hannel $\mathbf{S}$elf-$\mathbf{C}$lustering strategy, for linear models. Our Channel Self-Clustering (CSC) enhances CI strategy's performance improvements while reducing parameter size, for exmpale by over 10 times on electricity dataset, and significantly cutting training time. Second, we further propose Channel Rearrangement (CR), a method for deep models inspired by the self-clustering. CR attains competitive performance against baselines. Finally, we also discuss whether it is best to forecast the future values using the historical values of the same channel as inputs. We hope our findings and methods could inspire new solutions beyond CD/CI.
\end{abstract}

\section{Introduction}

Time series forecasting is the process of using past data to forecast future values, and it has gained increasing importance across various fields \citep{Khan2020TowardsEE, Angryk2020MultivariateTS, Chen2001FreewayPM}. Many model architectures, such as CNN-based \citep{Bai2018AnEE, Liu2021SCINetTS}, transformer-based \citep{Zhou2020InformerBE, Zhou2022FEDformerFE, Wu2021AutoformerDT} and linear models \citep{zeng2023transformers}, have been proposed and have outperformed traditional statistical methods.

Recent studies \citep{zeng2023transformers, Li2023RevisitingLT} have demonstrated that linear models are sufficiently effective in capturing temporal correlation compared to complex transformer-based models. Soon, PatchTST \citep{Nie2022ATS} has been proposed to demonstrate the effectiveness of transformers by introducing CI, partially inspired by linear models. \cite{Li2023RevisitingLT} have reported that CI can improve linear models on datasets with a large number of channels, enabling linear models to outperform PatchTST. Motivated by the advantages of CD/CI,  \cite{Chen2023MlinearRT} have enhanced linear models by employing CD and CI jointly. It appears that CI is currently one of the most effective strategies. In this paper, we investigate the CI from a question, whether linear models with CI benefits exactly from that using one individual layer for each channel. Then, we propose CSC and CR for linear and nonlinear models. Our methods do not require extensive parameters to obtain better performance than CI. Our contributions can be listed as
\begin{itemize}
    \item We investigate whether linear models with CI benefits from that using one individual layer for each channel by analyzing the generalization bound and empirically conclude that current CI for linear models leads to redundant layers.
    \item Our methods, CSC and CR for linear and nonlinear models, can adaptively determine the number of layers, achieving a trade-off between performance and model size while enhancing interpretability.
    \item We observe that it is not best to using the past values to forecast the future values of the same channel. It suggests the significant channel dependency yet to be exploited.
\end{itemize}

\section{Related Work}
A recent paper \citep{zeng2023transformers, Li2023RevisitingLT} has shown that simple linear models can outperform complicated transformer-based models \citep{Zhou2020InformerBE, Zhou2022FEDformerFE, Wu2021AutoformerDT, Liu2022PyraformerLP} and has cast doubt on the effectiveness of transformers on time series forecasting. PatchTST \citep{Nie2022ATS} has addressed this question by introduing Channel Dependence (CI) and patching into transformer. \cite{Han2023TheCA} have reported that CI can consistently enhance model performance. \cite{Li2023RevisitingLT} have studied the role of CI for linear models and demonstrated that linear models could be competitive on large datasets by using CI. Also, as reported in their work, CD is more effective on datasets with fewer channels. It might suggest that using an individual layer for each channel is unnecessary. MLinear \citep{Chen2023MlinearRT} is proposed to jointly use CD and CI to enhance the forecasting results. However, their method increases the model's size and complexity. The mechanism of CI deserves a fine-grained reconsideration. Rather than combining CD and CI, we investigate whether CI solely benefits from an extensive number of layers and propose CSC and CR to enhance linear and nonlinear models.

\section{Proposed Methods}

\subsection{Channel Self-Clustering}

Channel Self-Clustering employs the CI strategy as its initialization setup and pretrain it as a CI model during the first epoch. 

\paragraph{Channel Independent for linear models}

We denote a collection of $d$-dimensional multivariate time series samples with input horizon $L$ as an $\boldsymbol{X} \in \mathbb{R}^{d \times L}$. A linear model using CI modeling can be defined as $\theta_i, i \in (1, \ldots, d)$, which represents $d$ independent linear layers. We would like to forecast $T$ future values $\boldsymbol{Y} \in \mathbb{R}^{d \times T}$. The process of optimizing this model can be defined as 
\begin{align}
    \mathop{\arg\min}\limits_{\theta} \frac{1}{d} \sum_{i=1}^d \mathcal{L}\left(F\left(\boldsymbol{X}_i, \theta_i\right), \boldsymbol{Y}_i\right) \nonumber
\end{align}.


Subsequently, we validate each layer in this model using the validation set in a channel-dependent manner. Note that, the channel dependent strategy for linear models can be summarized as weight-sharing across channels. We calculate the error of each channel for each linear layer denoted by $\boldsymbol{E} \in \mathbb{R}^{d \times d}$, where $i$-th row $\boldsymbol{E}_i$ indicates the error of the $i$-th linear layer on all channels. We can deliberately select layer $\theta_{c_i}$ for $i$-th channel to minimize the validation loss:
\begin{align}
    \label{eq:SeC_layer_selection}
    c^* = \mathop{\arg\min}\limits_{c} \frac{1}{d} \sum_{i=1}^d \mathcal{L}\left(F\left(\boldsymbol{X}_i, \theta_{c_i}\right), \boldsymbol{Y}_i\right)
\end{align}
where $c \in \mathbb{Z}^{d}$. In the following epochs, we change our training objective into:
\begin{align}
    \label{eq:CSC_optimization}
    \mathop{\arg\min}\limits_{\theta} \frac{1}{d} \sum_{i=1}^d \mathcal{L}\left(F\left(\boldsymbol{X}_i, \theta_{c^*_i}\right), \boldsymbol{Y}_i\right) \nonumber
\end{align}.

In the following epochs, we can repeat the layer selection after training phases for several epochs.

Our method is inspired by the simple yet interesting fact that the learning process is inherently a self-clustering process that makes layers generalize for similar channels. 
The layer selection defined in Equation \ref{eq:SeC_layer_selection} can be explained as clustering, where we group together channels with similar periodic patterns and use a most generalizable layer to fit them. And we can analogize the forecasting error to the distance between data points and cluster centroids in a clustering algorithm. 

With a slight abuse of terminology, channel self-clustering implicitly trains layers of channels with a low signal-to-noise ratio (SNR) using data of high-SNR channels. In constrast, the CI strategy consistently trains a large number of layers on low-SNR channels, leading to compromised performance. Therefore, CSC can enhance the performance.


We empirically demonstrate that a large number of layers are pruned by the layer selection while improving the performance compared to CI. Moreover, the clusters of layers in CSC makes models more interpretable by explicitly showing clusters of a dataset.

\subsection{Generalization Bound of CSC}

We can decompose $i$-th channel $T^{(i)}$ into:
\begin{align}
    T^{(i)} = T^{(i)}_{SL} + T^{(i)}_{SU} + T^{(i)}_{NU} + T^{(i)}_{NM}
    \nonumber
\end{align}
where $T^{(i)}_{SL}$, $T^{(i)}_{SU}$, $T^{(i)}_{NU}$  and $T^{(i)}_{NM}$ denote the learnable signal component, the unlearnable signal component, the memorizable noise component and the unmemorizable noise component respectively. Since we only discuss under the condition of not changing the model capacity, we can denote $T^{(i)}_{SU} + T^{(i)}_{NU} + T^{(i)}_{NM}$ as broadly defined noise $N^{(i)}$ and $T^{(i)}_{SL}$ as signal $S^{(i)}$. Then, we can denote the input matrix and the target matrix as $X^{(i)}= X^{(i)}_S + X^{(i)}_N$ and $Y^{(i)}= Y^{(i)}_S + Y^{(i)}_N$. We define a well-trained model $\mathcal{H}^*$ trained on training set $\hat{T}^{(i)}$ as a signal error minimizer such that $\mathcal{H}^*\left(X^{(i)}_S\right) = Y^{(i)}_S$. Note that $\mathcal{H}^*$ is not necessarily the generalization error minimizer.

For a linear-CSC model, if $c^*_i$ is not equal to $i$, we can reasonably assume the features in $X^{(i)}_S$ are predominantly shared with $X^{(c^*_i)}_S$. For the convenience of discussion, we assume the signal error minimizer $\mathcal{H}^*_{(c^*_i)}$ for the $c^*_i$-th channel is also the minimizer for the $i$-th channel. Then, we have
\begin{align}
    \mathcal{L}_{T^{(i)}}\left(\mathcal{H}^*_{(c^*_i)}\right) &= \mathbb{E}\left[\mathcal{L}\left(\mathcal{H}^*_{(c^*_i)}\left(x^{(i)}_N\right), y^{(i)}_N\right)\right] \nonumber
\end{align}
. And we assume that the weights of $\theta_i$ and $\theta_{c^*_i}$ have similar $l_2$ norm.

\begin{lemma}
    Let $\mathcal{H} = \{x \rightarrow \left\langle w, x\right\rangle | w \in \mathbb{R}^d, \left\|w\right\|_2 \leq B \}$ for some constant $B \geq 0$, Moreover, assume each element of $x$ is a random variable independently drawn from a $\sigma$-sub-Gaussian distribution. Then, with probability at least $1 - \delta$, the upper bound of the Rademacher complexity of $\mathcal{H}$ is
    \begin{align}
    R_n\left(\mathcal{H}\right) &\leq BC\sigma\sqrt{\frac{d}{n}} \nonumber \\
    C &= \sqrt{2 \log{\frac{2}{\delta}}} \nonumber
    \end{align}
\end{lemma}

\begin{lemma}
    Let $X_1, \ldots, X_n$ be independent sub-Gaussian random variables with variance proxies $\sigma^2_1, \ldots, \sigma^2_n$. Define $Z = \left\langle \mathbf{o}, [X_1, \ldots, X_n] \right\rangle $ where $\mathbf{o}$ is a random one-hot vector following a uniform distribution over all one-hot vectors. Then, we have
    \begin{align}
    \mathrm{P}\left(\lvert Z - \mathbb{E}[Z] \rvert \geq t \right) &\leq 2 \exp\left(-\frac{t^2}{2 \max\{\sigma^2_1, \ldots, \sigma^2_n\}}\right) \nonumber
    \end{align}
\end{lemma}

\begin{corollary}
    \label{cor:linear_rad_complexity}
    Let $\mathcal{H}^*_{CSC}$ be a well-trained linear-CSC model with one layer dominating $K$ channels and $\mathcal{H}^*_{CI}$ a well-trained linear-CI model trained on the same $K$ channels. Assume $N^{(i)}$ is a sub-Gaussian random variable with variance proxy $\sigma^2_i$. We have
    \begin{align}
    \mathcal{L}\left(\mathcal{H}^*_{CSC}\right) - \hat{\mathcal{L}}\left(\mathcal{H}^*_{CSC}\right)
     &\leq R_{CSC} + \epsilon_{CSC}
    \nonumber \\
    \mathcal{L}\left(\mathcal{H}^*_{CI}\right) - \hat{\mathcal{L}}\left(\mathcal{H}^*_{CI}\right)
    &\leq R_{CI} + \epsilon_{CI} \nonumber
    \end{align}
    With probability at least $1 - \lambda$, 
    \begin{align}
    R_{CSC} &= \mathcal{O}\left(\frac{\max \{\sigma_1, \sigma_2, \ldots, \sigma_K\}}{\sqrt{Kn}}\right) \nonumber \\ 
    R_{CI} &= \mathcal{O}\left(\frac{\sum^K_{i=1}{\sigma_i}}{K\sqrt{n}}\right) \nonumber
    \end{align}
    With probability at least $1 - \delta$, 
    \begin{align}
    \epsilon_{CSC} = \mathcal{O}\left(\sqrt{\frac{\ln{\left(\frac{1}{\delta}\right)}}{Kn}}\right) \nonumber \\ 
    \epsilon_{CI} = \mathcal{O}\left(\sqrt{\frac{\ln{\left(\frac{1}{\delta}\right)}}{n}}\right) \nonumber
    \end{align}
\end{corollary}

Corollary \ref{cor:linear_rad_complexity} demonstrates that the generalization bound of linear-CSC models is more likely to be smaller than linear-CI models, particularly when $K$ is large, since the noise variance of similar channels typically do not differ significantly.

\subsection{Channel Rearrangement}

We assume a deep nonlinear time series forecasting model contains an encoder $E$ and a decoder $D$. For example, in a 2-layer MLP, we refer to the bottommost linear layer as the encoder $E$.

To achieve accurate channel clustering, we pretrain the model as a whole for several epochs, since deep models are prone to underfitting when trained with insufficient channels and training steps. The training process can be defined as
\begin{align}
    \mathop{\arg\min}\limits_{E, D} \frac{1}{d} \sum_{i=1}^d \mathcal{L}\left(F\left(\boldsymbol{X}_i, E, D\right), \boldsymbol{Y}_i\right) \nonumber
\end{align}

After pretraining, we replicate the encoder $d$ times where $d$ is the number of channels. Each channel is assigned to one of the encoder replicates. We optimize the encoders with the following objective
\begin{align}
    \mathop{\arg\min}\limits_{E} \frac{1}{d} \sum_{i=1}^d \mathcal{L}\left(F\left(\boldsymbol{X}_i, E_i, D\right), \boldsymbol{Y}_i\right) \nonumber
\end{align}
where $E_i$ denotes the $i$-th encoder for the $i$-th channel. 

Then, we update $c$ to be a mapping vector that minimizes the loss:
\begin{align}
    c^* = \mathop{\arg\min}\limits_{c} \frac{1}{d} \sum_{i=1}^d \mathcal{L}\left(F\left(\boldsymbol{X}_i, E_{c_i}, D\right), \boldsymbol{Y}_i\right) \nonumber
\end{align}
In the following epochs, we change our training objective into:
\begin{align}
    \mathop{\arg\min}\limits_{E} \frac{1}{d} \sum_{i=1}^d \mathcal{L}\left(F\left(\boldsymbol{X}_i, E_{c^*_i}, D\right), \boldsymbol{Y}_i\right) \nonumber
\end{align}.

Similarly, we can iteratively update the mapping vector after training phases for several epochs. Updating the mapping vector is equivalent to rearranging each channel to a new encoder.

Practitioners who want to follow this method can repeat and assgin other modules, such as the decoder, to reduce computational cost and it maintains similar performance.

CR implicitly introduces the channel information and thus enhance the forecasting performance. In contrast, we do not observe noticeable improvement caused by explicitly adding the channel information using channel embedding \citep{Shao2022SpatialTemporalIA, lin2023segrnn}.

\section{Experiments}

\subsection{Channel Self-Clustering}

We benchmark RLinear \citep{Li2023RevisitingLT} using CSC modeling against CD and CI modeling on 5 real-world datasets, following the experimental setup of previous work. RLinear is a linear model wrapped by reversible instance normalization \citep{Kim2022ReversibleIN}. Table 2 provides statistical information of those 5 real-world datasets. For RLinear-CSC, we perform the layer selection in the first two epochs. We train all these models for 10 epochs including the epochs of the layer selecion.

We compare performance using Mean Squared Error (MSE) and Mean Absolute Error (MAE) as the core metrics. And we also compare the number of model parameters.

As shown in Table \ref{tab:CSC_performance},  CSC can improve the performance compared to CI while significantly reducing the model size compared to CI. CSC also significantly reduces the training time. For example, training RLinear-CI on ECL dataset with forecasting horizon 720 costs more than 2800 seconds while RLinear-CSC, according to our implementation, only takes less than 1200 seconds on the same device. In fact, we can obtain better performance by adjusting the number of epochs for the layer selection on different datasets. Reducing the number of epochs for the layer selection is preferable for high-SNR datasets, such as weather. Our experimental results demonstrate that it is not optimal to use one individual layer to model each channel.

\begin{table*}[htb]
    \centering
    \caption{Statistics of all datasets for experiments.}
    \label{tab:datasets}
    \label{tab:2}
    \vskip 0.15in
    \setlength\tabcolsep{14pt}
    \begin{small}
    \begin{tabular}{c|cc|ccc}
    \toprule
    Dataset & ETTm1 & ETTm2 & Weather & ECL & Traffic \\
    \midrule
    N Channels & 7 & 7 & 21 & 321 & 862 \\
    Timesteps & 69,680 & 69,680 & 52,696 & 26,304 & 17,544 \\
    Granularity & 15 min & 15 min & 10 min & 1 hour & 1 hour \\
    \midrule
    Data Partition & \multicolumn{2}{c|}{6:2:2} & \multicolumn{3}{c}{7:2:1} \\
    \bottomrule
    \end{tabular}
    \end{small}
    \vskip -0.1in
\end{table*}

\begin{table*}[!htb]
    \centering
    \caption{Time series forecasting results. The length of the historical horizon is 336 and prediction lengths are \{96, 192, 336, 720\}. The best results are in \textbf{bold} and the second one is \underline{underlined}. RMP denotes the ratio of model parameters between final CSC and CI. The number inside the (parentheses) represents the number of layers preserved after the layer selection.}
    \label{tab:CSC_performance}
    \vskip 0.15in
    \setlength\tabcolsep{6.3pt}
    \begin{small}
    \begin{tabular}{c|c|ccccccc}
    \toprule
    \multicolumn{2}{c|}{Model} & \multicolumn{7}{c}{RLinear} \\
    \midrule
    \multicolumn{2}{c|}{Method} & \multicolumn{2}{c|}{CD} & \multicolumn{2}{c|}{CI} & \multicolumn{3}{c}{CSC} \\
    \midrule
    \multicolumn{2}{c|}{Metric} & MSE & MAE & MSE & MAE & MSE & MAE & RMP \\
    \midrule
    \multirow{4}{*}{\rotatebox{90}{ETTm1}}
     & 96 & 0.306 & 0.347 & \underline{0.292} & 0.339 & \textbf{0.291} & \textbf{0.338} & 42.8 \% (3) \\
    & 192 & 0.343 & 0.369 & \textbf{0.333} & \textbf{0.362} & \underline{0.338} & \underline{0.366} & 42.8 \% (3) \\
    & 336 & \underline{0.373} & \underline{0.385} & 0.376 & 0.387 & \textbf{0.371} & \textbf{0.383} & 42.8 \% (3) \\
    & 720 & \textbf{0.426} & \textbf{0.415} & 0.430 & \underline{0.416} & \underline{0.429} & \underline{0.416} & 57.1 \% (4) \\
    \midrule
    \multirow{4}{*}{\rotatebox{90}{ETTm2}}
     & 96 & \underline{0.166} & \underline{0.256} & \textbf{0.165} & \textbf{0.253} & \textbf{0.165} & \textbf{0.253} & 42.8 \% (3) \\
    & 192 & 0.222 & 0.293 & \underline{0.220} & \underline{0.291} & \textbf{0.219} & \textbf{0.290} & 57.1 \% (4) \\
    & 336 & \underline{0.277} & \underline{0.329} & \textbf{0.275} & \textbf{0.327} & \underline{0.277} & 0.330 & 57.1 \% (4) \\
    & 720 & \textbf{0.370} & \underline{0.388} & \underline{0.371} & \textbf{0.387} & \underline{0.371} & 0.389  & 42.8 \% (3) \\
    \midrule
    \multirow{4}{*}{\rotatebox{90}{Traffic}}
     & 96 & \underline{0.410} & \textbf{0.278} & 0.422 & 0.292 & \textbf{0.409} & \textbf{0.278} & 1.5 \% (13) \\
    & 192 & \textbf{0.422} & \textbf{0.282} & \underline{0.433} & \underline{0.295} & \textbf{0.422}  & \textbf{0.282}  & 1.6 \% (14) \\
    & 336 & \textbf{0.435} & \textbf{0.289} & 0.447 & 0.303 & \underline{0.438} & \underline{0.290} & 2.6 \% (23) \\
    & 720 & \textbf{0.463} & \textbf{0.306} & 0.472 & 0.317 & \underline{0.465} & \underline{0.308} & 2.9 \% (25) \\
    \midrule
    \multirow{4}{*}{\rotatebox{90}{Weather}}
     & 96 & \underline{0.175} & 0.224 & \textbf{0.145} & \underline{0.194} & \textbf{0.145} & \textbf{0.193} & 52.3 \% (11) \\
    & 192 & 0.217 & 0.259 & \underline{0.189} & \underline{0.235} & \textbf{0.188} & \textbf{0.234} & 47.6 \% (10) \\
    & 336 & 0.264 & 0.293& \underline{0.242} & \underline{0.275} & \textbf{0.240} & \textbf{0.274} & 52.3 \% (11) \\ 
    & 720 & 0.328 & 0.339 & \underline{0.314} & \underline{0.327} & \textbf{0.312} & \textbf{0.325} & 42.8 \% (9) \\
    \midrule
    \multirow{4}{*}{\rotatebox{90}{ECL}}
     & 96 & \underline{0.140} & \underline{0.235} & \textbf{0.134} & \textbf{0.228} & \textbf{0.134} & \textbf{0.228} & 9.3 \% (30) \\
    & 192 & \underline{0.153} & \underline{0.247} & \textbf{0.149} & \textbf{0.242} & \textbf{0.149} & \textbf{0.242} & 4.9 \% (16) \\
    & 336 & 0.170 & 0.263 & \underline{0.166} & \underline{0.259} & \textbf{0.165} & \textbf{0.258} & 8.4 \% (27) \\
    & 720 & 0.209 & \underline{0.296} & \underline{0.205} & \textbf{0.291} & \textbf{0.204} & \textbf{0.291} & 6.8 \% (22) \\
    \bottomrule
    \end{tabular}
    \end{small}
\end{table*}

\begin{figure}[htbp]
    \centerline{\includegraphics[scale=0.6]{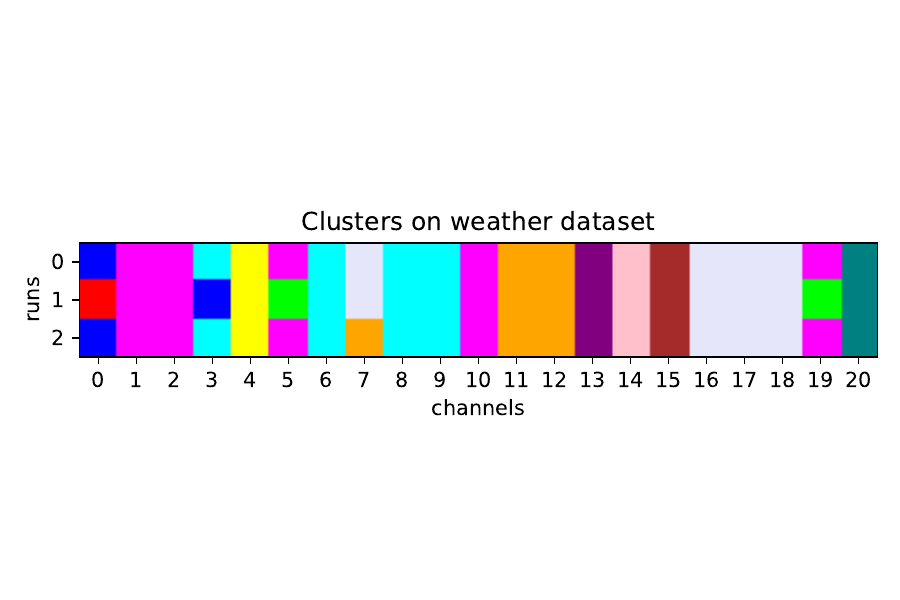}}
    \caption{Channels dominated by the same layer are represented in the same color. The figure displays the clusters resulting from 3 runs on weather dataset.}
    \label{fig:wth_clusters}
\end{figure}

We also train RLinear using CSC with different random seeds. As illustrated in figure \ref{fig:wth_clusters}, we find that clusters of channels are relatively stable. It indicates CSC does group channels with similar inherent features rather than randomly grouping. The clusters of channels enhance the model interpretability.

\subsection{Channel Rearrangement}

We run this experiment on three datasets, weather, ECL and traffic, since MLP is well-suited for large datasets.

In this experiment, we benchmark MLP-CR against MLP-CI and MLP-CE where the output of the encoder of each channel is concatenated with a unique channel embedding \citep{Shao2022SpatialTemporalIA, lin2023segrnn}.
We train models for 18 epochs with same setup. MLP-CR is pretrained for 15 epochs and updates its mapping vector in last the 3 epochs. Similarly, MLPs are wrapped by reversible instance normalization and the activation function used here is Swish \citep{Elfwing2017SigmoidWeightedLU, Ramachandran2017SwishAS}.


For a fair comparison, we additionally train all baseline models using merged sets which consist of both the training set and validation set. By doing this, we ensure that the improvement from CR is not due to the information leakage from the validation set.



As illustrated in Table \ref{tab:CR_performance}, CR contributes a clear improvement compared to baseline models. Notably, we do not observe a significant performance improvement with the channel embedding in this experiment.

\begin{table*}[!htb]
    \centering
    \caption{Time series forecasting results. The length of the historical horizon is 336 and prediction lengths are \{96, 192, 336, 720\}. The best results are in \textbf{bold}. The results inside the (parentheses) are trained using merged datasets.}
    \label{tab:CR_performance}
    \vskip 0.15in
    \setlength\tabcolsep{6.3pt}
    \begin{small}
    \begin{tabular}{c|c|cccccc}
    \toprule
    \multicolumn{2}{c|}{Model} & \multicolumn{6}{c}{MLP} \\
    \midrule
    \multicolumn{2}{c|}{Method} & \multicolumn{2}{c|}{CI} & \multicolumn{2}{c|}{CE} & \multicolumn{2}{c}{CR} \\
    \midrule
    \multicolumn{2}{c|}{Metric / Hyperparam} & MSE & MAE & MSE & MAE & MSE & MAE \\
    \midrule
    \multirow{4}{*}{\rotatebox{90}{Weather}}
     & 96 & 0.147 (0.147) & 0.198 (0.198) & 0.147 (0.147) & \textbf{0.197} (0.198) & \textbf{0.146} & \textbf{0.197} \\
    & 192 & 0.190 (0.190) & 0.240 (0.240) & 0.190 (0.190) & 0.239 (\textbf{0.239}) & \textbf{0.188} & \textbf{0.239} \\
    & 336 & 0.244 (0.242) & 0.280 (0.279) & 0.243 (0.243) & 0.279 (\textbf{0.278}) & \textbf{0.240} & \textbf{0.278} \\
    & 720 & 0.317 (\textbf{0.316}) & 0.332 (\textbf{0.329}) & 0.318 (0.318) & 0.331 (0.331) & \textbf{0.316} & 0.330 \\
    \midrule
    \multirow{4}{*}{\rotatebox{90}{ECL}}
     & 96 & 0.130 (0.130) & 0.224 (0.224) & 0.130 (0.130) & 0.224 (0.224) & \textbf{0.129} & \textbf{0.222} \\
    & 192 & 0.148 (\textbf{0.147}) & 0.240 (\textbf{0.239}) & 0.148 (0.148) & 0.240 (0.240) & \textbf{0.147} & \textbf{0.239} \\
    & 336 & 0.163 (\textbf{0.163}) & 0.257 (\textbf{0.256}) & 0.164 (0.165) & 0.257 (0.257) & \textbf{0.163} & \textbf{0.256} \\
    & 720 & 0.203 (0.202) & 0.291 (0.290) & 0.205 (0.201) & 0.295 (0.290) & \textbf{0.200} & \textbf{0.289} \\
    \midrule
    \multirow{4}{*}{\rotatebox{90}{Traffic}}
     & 96 & 0.375 (0.382) & 0.262 (0.262) & 0.377 (0.382) & 0.263 (0.263) & \textbf{0.371} & \textbf{0.261} \\
    & 192 & 0.395 (0.403) & 0.271 (\textbf{0.270}) & 0.396 (0.402) & 0.271 (0.271) & \textbf{0.392} & \textbf{0.270} \\
    & 336 & 0.410 (0.432) & \textbf{0.278} (\textbf{0.278}) & 0.410 (0.430) & 0.279 (\textbf{0.278}) & \textbf{0.407} & \textbf{0.278} \\
    & 720 & 0.441 (0.507) & \textbf{0.295} (0.297) & 0.441 (0.493) & 0.297 (0.297) & \textbf{0.440} & 0.296 \\
    \bottomrule
    \end{tabular}
    \end{small}
\end{table*}

\subsection{Whether it is best to forecast future values only using the past values of the same channel?}
\label{discuss}

Using past values to forecast this channel is the core characteristic of CI strategy. However, we wonder whether it is the best strategy. To answer this, we train a linear model with $d \times d$ layers $L$ where $d$ is the number of channels and the layer $L_{i, j}$ is trained to use the $i$-th channel to forecast the $j$-th channel.

\begin{figure*}[htbp]
    \centering
    \centering
    \includegraphics[scale=0.42]{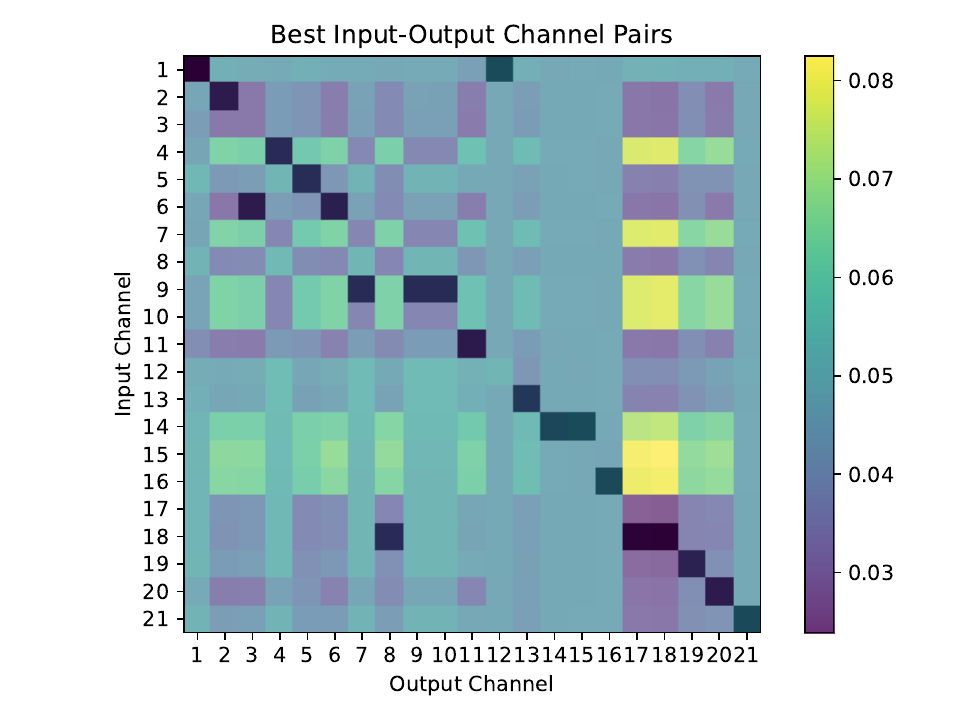}
    \label{fig:sub1}
    \centering
    \includegraphics[scale=0.42]{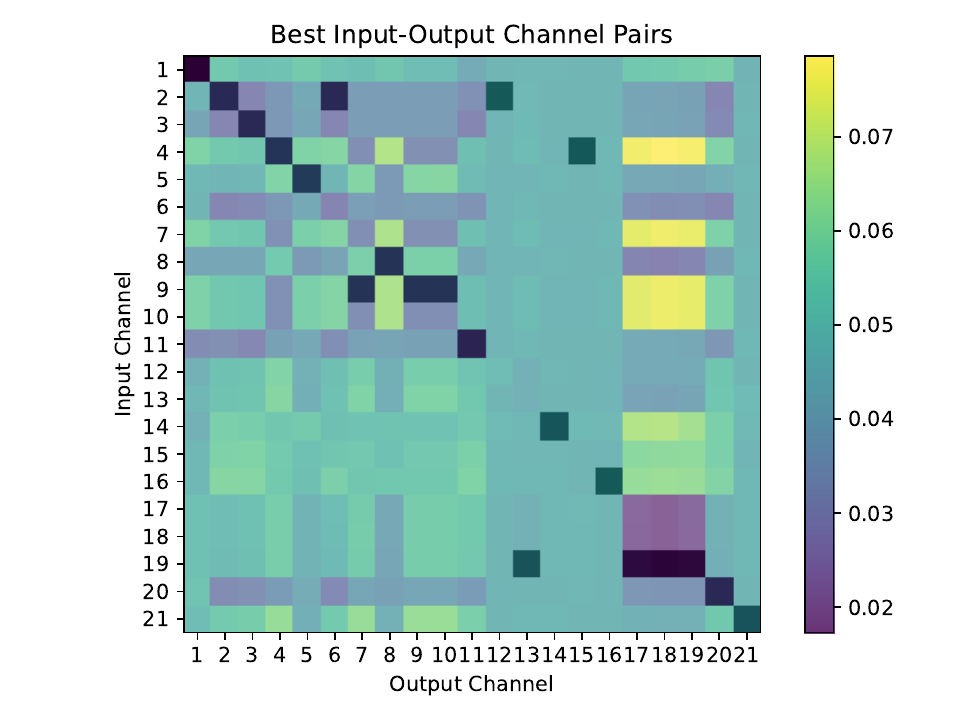}
    \label{fig:sub2}
    \centering
    \centering
    \includegraphics[scale=0.45]{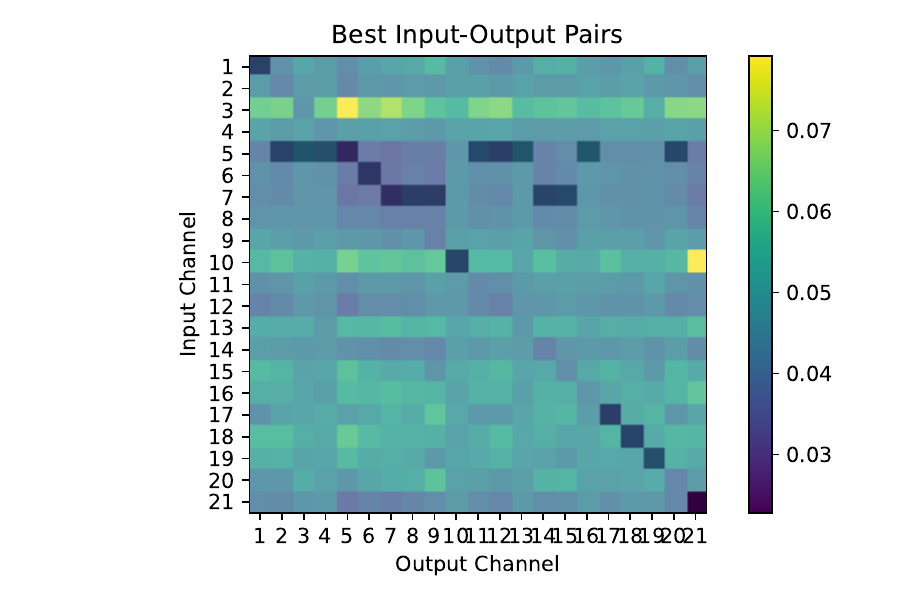}
    \label{fig:sub3}
    \centering
    \includegraphics[scale=0.45]{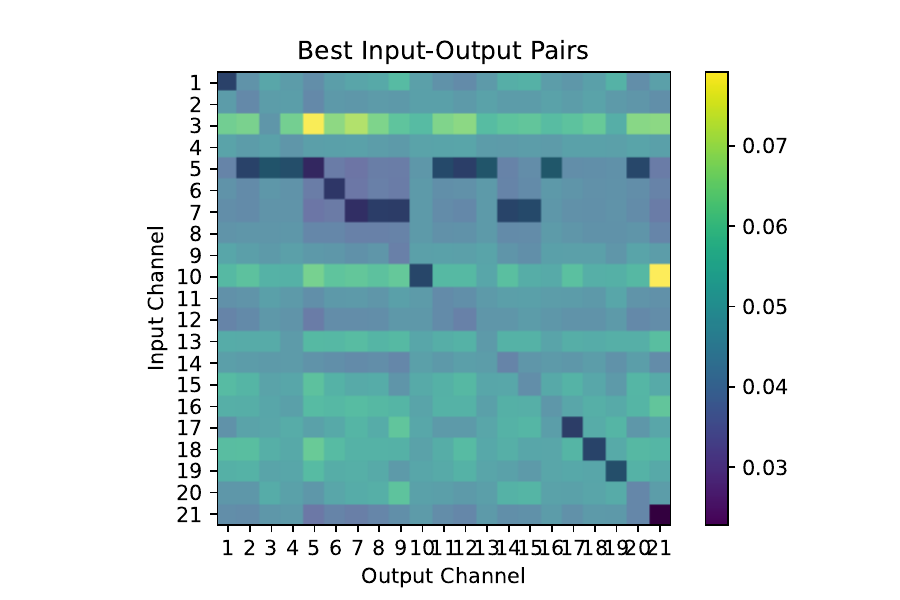}
    \label{fig:sub4}
    \caption{We visualize the normalized loss matrices on different datasets. \textbf{upper left}: the training set of weather dataset; \textbf{ upper right} : the test set of weather dataset; \textbf{lower left}: the first 21 channels of the training set of traffic dataset; \textbf{lower right}: the first 21 channels of the test set of traffic dataset. The darkest cell in each column indicates the input channel with the lowest loss.}
    \label{fig:cross_channel}
\end{figure*}

Surprisingly, we find that $L_{i, i}$ does not always result in the lowest loss. As shown in Figure \ref{fig:cross_channel}, some other channels might be more useful than the forecasting channel. For example, to forecast the 7-th channel on weather dataset, the best input is the 9-th channel. 

This phenomenon indicates CI which simply uses the current channel as inputs should be revisited and more effective channel mixing methods are yet to be explored.

\section{Conclusion and Future Work}

This paper rethinks the channel independent strategy from two perspectives and concludes that (1) it is unnecessary to use one layer to model each channel in the linear CI strategy and (2) it is not best to forecast the future values using the same channel as the inputs. As a result, we propose two simple yet effective methods called CSC and CR for linear and nonlinear models respectively. CSC is proven to be able to reduce the computational cost and model size while improving performance. CR can enhance the forecasting results for nonlinear.

As discussed in section \ref{discuss}, cross-channel dependency is yet to be exploited and it would be an important next step.

\bibliography{iclr2023_conference}
\bibliographystyle{iclr2023_conference}

\end{document}